%% file: main.tex
\documentclass[10pt,twocolumn,letterpaper]{article}

\usepackage[pagenumbers]{wacv}              %

\input{preamble}
\definecolor{wacvblue}{rgb}{0.21,0.49,0.74}
\usepackage[,pagebackref,breaklinks,colorlinks,allcolors=wacvblue]{hyperref}
\usepackage{multirow}

\title{Beyond Contact Sensors: Deep learning with Pseudo-Labeling for remote Photoplethysmography}

\author{Bhargav Acharya, Barbara Hammer, Hanna Drimalla\\
Center for Cognitive Interaction Technology (CITEC), Bielefeld University\\
{\tt\small \{bacharya, bhammer, drimalla\}@techfak.uni-bielefeld.de}
}

\begin{document}

\maketitle
\input{sections/0_abstract}
\input{sections/1_intro}

\input{sections/2_related_works}

\input{sections/3_methods}

\input{sections/4_experiments}
\input{sections/5_results}

\input{sections/6_discussions}

\input{sections/7_conclusion}
{
	\small
	\bibliographystyle{ieeenat_fullname}
	\bibliography{main}
}

\end{document}

%% file: sections/0_abstract.tex
\begin{abstract}
	Heart rate is a critical biomarker of health, and remote photoplethysmography (rPPG) enables its contactless estimation from video data for telemedicine applications. Recent advancements in deep learning based rPPG methods achieve state-of-the-art results, outperforming classical signal-processing methods in complex scenarios. However, deep learning methods depend on datasets with precise synchronization between videos and ground truth signals collected via contact sensors, whereas signal-processing-based methods do not. To address this dependence on labeled datasets, which are labor-intensive to collect, we investigate under which circumstances pseudo-labels extracted using unsupervised signal-processing methods can replace contact sensors labels for training deep learning methods. Our systematic evaluations found that for datasets with imperfect synchronization, the pseudo-label approach outperforms supervised training on contact sensors. For datasets with good synchronization, results are mixed: within-dataset evaluation shows no significant difference between training methods, while cross-dataset evaluation favors supervised training. However, removing a single outlier participant significantly improves the pseudo-label approach's cross-dataset performance, highlighting the importance of label quality. These results demonstrate that signal-processing methods can generate valid training signals for deep learning models, reducing dependency on labor-intensive dataset collection while maintaining competitive performance.
\end{abstract}

\makeatletter
\renewcommand{\@makefnmark}{}  %
\renewcommand{\@makefntext}[1]{\noindent#1}  %
\makeatother
\footnotetext{This work was supported by SAIL, funded by the Ministry of Culture and Science of the State of North Rhine-Westphalia under the grant no NW21-059A}

%% file: sections/1_intro.tex
\section{Introduction}

Remote photoplethysmography (rPPG) is a non-invasive technique that measures
subtle blood volume changes from videos of the face, enabling estimation of vital
physiological parameters like heart rate and respiration \cite{Xiao2024}.
The non-invasive nature and lack of dedicated hardware requirements make rPPG highly relevant for telehealth and
remote monitoring applications.

rPPG methods can be broadly classified into two categories: signal-processing-based
(sometimes called conventional) and supervised deep learning based methods \cite{Xiao2024}.
Signal-processing approaches rely on predetermined physiological
models and handcrafted features to extract signals from videos and are inherently
unsupervised. In contrast, deep learning methods learn to predict physiological
signals directly from data, often outperforming signal-processing techniques and
achieving state-of-the-art results on public datasets \cite{Xiao2024}. However,
the supervised nature of deep learning based methods requires large annotated
datasets consisting of videos with synchronized photoplethysmography (PPG) ground
truth from contact devices, making data collection labor-intensive and costly.

Existing datasets are limited and typically represent idealistic scenarios with minimal movement, good lighting, and limited heart rate ranges \cite{Heusch2017,Xiao2024}.
This leads to deep learning models that overfit to training data and show significant performance degradation in real-world applications \cite{Dasari2021d,Acharya2025}.
Furthermore, majority of datasets use finger-mounted contact devices to collect PPG ground truth data for supervised training.
However, PPG signals from different body sites exhibit distinct morphological characteristics and temporal phase \cite{Niu2023}.
This leads to a mismatch between the signal that is being extracted from the videos and the ground truth signal-provided for supervised training.
In this regard, recent work by \cite{Braun2024a} showed that training the deep learning methods on contact sensors placed at the face leads to better performing deep learning models.
An additional source of mismatch stems from synchronization quality during dataset
collection. Videos and contact sensors are recorded using independent devices, requiring
dedicated hardware or software to achieve precise alignment.
To the best of our knowledge, only the PURE dataset \cite{Stricker2014a} employs external hardware to
achieve near-perfect synchronization whereas other datasets use software-based synchronizations.
This lack of perfect synchronization, combined with the physiological mismatch
between finger-based ground truth and face-extracted signals, have
prevented deep learning methods from reaching their full potential on existing public
datasets.

To bridge this domain gap, new datasets with face-mounted contact sensors would be
needed to eliminate the site mismatch. However, collecting such datasets is
labor-intensive, and face-mounted sensors obscure facial features and limit
realistic video recording.

Addressing these limitations requires alternate training strategies that reduce the
dependence on datasets with finger contact sensors without sacrificing model performance.
To this end, we evaluate the feasibility of using pseudo-labels which can be automatically extracted
using signal-processing-based rPPG methods, specifically POS \cite{Wang2017},
for training deep learning models in a weakly supervised manner, thereby presenting an opportunity to replace contact sensor labels.

Our contributions are as follows,
we systematically compare deep learning based rPPG
methods trained on three different signals: contact-based ground truth (Finger-PPG),
signal-processing-based pseudo-labels extracted from videos (Pseudo-PPG), and
temporally aligned ground truth (Synced-PPG). We perform within-dataset and
cross-dataset experiments to evaluate whether training signal choice significantly
affects model performance.

%% file: sections/2_related_works.tex
\section{Related Work}

\subsection{rPPG Methods}
rPPG has received significant attention in recent years, progressing from signal-processing approaches to deep learning methods.
Early rPPG research focused on signal-processing methods that extract physiological signals without supervised learning.
GREEN \cite{Verkruysse2008} showed that the green channel best captures pulse signals through hemoglobin absorption.
Subsequent methods employed blind source separation techniques to decompose RGB signals into independent components \cite{Poh2010,Madej2011}.
Other methods, including POS \cite{Wang2017} and CHROM \cite{deHaan2013c}, used skin reflectance properties to extract rPPG signals.

Supervised deep learning methods now dominate rPPG research and have consistently achieved state-of-the-art performance on publicly available datasets
\cite{Liu2023b}.
DeepPhys \cite{Chen2018c}, TS-CAN \cite{Liu2020}, and Physnet \cite{Yu2019e} have become established benchmarks against which
new architectures are evaluated, demonstrating that end-to-end learning can outperform traditional signal-processing approaches.
Recent transformer-based methods like PhysFormer \cite{Yu2022b}, EfficientPhys \cite{Liu2022b}, RhythmFormer \cite{Zou2025}, and PhysMamba \cite{Luo2024}
further improved accuracy through advanced temporal modeling.

Despite these advances, all supervised methods require synchronized contact PPG
sensors during training, creating a scalability bottleneck.
While signal-processing-based methods are computationally efficient and require no training, they typically exhibit
higher error rates than supervised deep learning approaches in complex scenarios \cite{Liu2023b}.

\subsection{Training Signal}

Most deep learning methods use finger-based contact PPG sensors as training labels \cite{Xiao2024}.
While PPG signals are known to differ across body sites due to pulse transit time (PTT) and morphological variations \cite{Lindholm2007},
the impact on deep learning performance remained unclear until recently.
\cite{Braun2024a} demonstrated that training with forehead-mounted
contact PPG reduces mean squared error between predicted and ground truth waveforms
by up to 40\% compared to finger PPG, attributing this improvement to reduced domain
gap when input (facial videos) and labels (facial PPG) originate from the same region.

\subsection{Weakly Supervised Learning}
Weakly supervised learning addresses the challenge of obtaining high-quality labels by training models with cheaper, noisier supervision.
\citet{Lee2013} first introduced pseudo-labeling for deep neural networks,
where model predictions on unlabeled data serve as surrogate training labels.
This technique has now been applied across various computer vision tasks \cite{Kage2025}.

In rPPG research, training has predominantly required synchronized contact sensors for ground truth.
Recent work has begun reducing this dependency through contrastive
learning frameworks \cite{Gideon2021a}. \citet{Li2023a} extended this by
incorporating pseudo-labels from the 2SR algorithm \cite{Wang2016} within their
contrastive pipeline, using losses from both predicted signals and pseudo-labels
to guide self-supervised learning. Similarly, \cite{Wu2025b} explored
use of pseudo-labels in a self-supervised setting by training a deep learning model
for one epoch on labeled data, then using this model to generate pseudo-labels for
unlabeled data to expand the training set.

Althoug these prior studies incorporated pseudo-labels, their use as the sole training
signal has not yet been systematically evaluated. Our work contributes in this direction in two key aspects.
First, we systematically investigate whether
signal-processing algorithms can directly generate pseudo-labels for training deep learning methods,
without requiring an initial labelled dataset.
Second, we train the models using pseudo-labels with mean squared error loss,
eliminating the need for contrastive learning frameworks or specialized loss functions.
This provides a direct assessment of whether signal-processing based training labels
can serve as a viable alternative to contact sensor labels.

\subsection{Pseudo-PPG}
To the best of our knowledge, only two prior works have used POS \cite{Wang2017}, a signal-processing based
method, for generating training labels \cite{Zhan2020,Liu2023b}.
\citet{Zhan2020} investigated whether physiological delays
caused by PTT critically affect deep learning training. They used POS-extracted
signals as training labels instead of finger PPG to eliminate temporal misalignment,
demonstrating through synthetic phase shift experiments that alignment between
training labels and input videos is critical for effective training. rPPG-Toolbox
\cite{Liu2023b} also uses POS-generated signals as a practical workaround when
high-quality synchronized labels are unavailable.
However, these works either use pseudo-labels for signal alignment or as a fallback
strategy when synchronized data is unavailable, rather than systematically evaluating
whether POS-extracted signals can replace contact sensor ground truth as primary
training signals for deep learning methods.

Our work directly investigates whether POS-extracted signals can be treated
as primary training labels.
If models trained on Pseudo-PPG achieve comparable performance to those trained
on contact PPG, the dependence on synchronized finger sensors is reduced,
lowering dataset acquisition costs and deployment barriers.

%% file: sections/3_methods.tex
\section{Methods}

This section describes the deep learning architectures used for
rPPG estimation, the training signals employed, and the complete rPPG processing
pipeline.

\subsection{Deep Learning Methods}
\label{sec:deep-learning-methods}
We evaluate two benchmark deep learning based rPPG methods: TS-CAN \cite{Liu2020}
and Physnet \cite{Yu2019e}.
These well-established architectures use CNN-based designs
rather than large transformer models, enabling us to isolate training signal effects
from architectural innovations.

TS-CAN uses a dual-branch architecture with temporal
shift modules for temporal modeling and 2D convolutions for spatial feature
extraction, combined through an attention mechanism.

The temporal shift branch receives diff-normalized input computed from adjacent
frames:
\begin{equation}
	\label{eq:diffnorm}
	c_{\text{diff}}(t) = \frac{c(t+1) - c(t)}{c(t+1) + c(t)}
\end{equation}
where $c(t)$ denotes the video frame at time $t$. The appearance branch receives
standardized input:
\begin{equation}
	c_{\text{std}}(t) = \frac{c(t) - \mu}{\sigma}
\end{equation}
where $\mu$ and $\sigma$ are the mean and standard deviation computed across
each video chunk.

Physnet employs a 3D convolutional encoder-decoder architecture to jointly
process spatial and temporal dimensions. Input frames are diff-normalized using
the same procedure as the TS-CAN temporal branch as described in \autoref{eq:diffnorm}.

\subsection{Training Signals}
\label{sec:training-signals}
\begin{figure}[ht]
	\centering
	\includegraphics[width=\columnwidth]{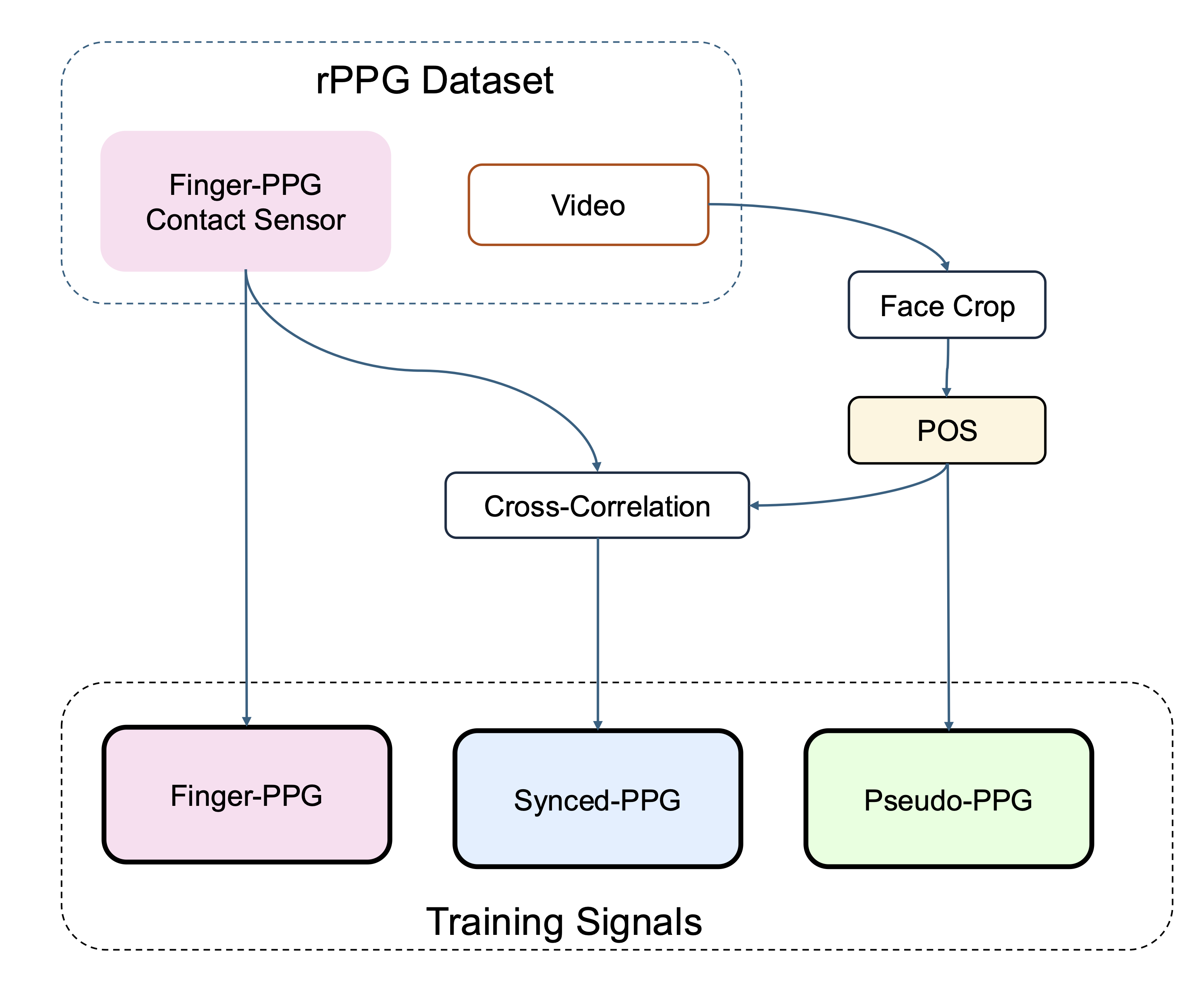}
	\caption{ \textbf{Generating the training signals from rPPG dataset recordings.} Finger-PPG is taken directly from the contact sensor as the gold-standard reference.
		Pseudo-PPG is extracted from facial video via face cropping and the POS \cite{Wang2017}.
		Synced-PPG is obtained by temporally aligning the finger-PPG signal to the POS-derived facial signal via cross-correlation.
	}
	\label{fig:training-signals}
\end{figure}

To evaluate whether signal-processing based pseudo-signals can replace contact-based ground
truth, we train models using three different signals: Finger-PPG, Pseudo-PPG,
and Synced-PPG. Both Pseudo-PPG and Synced-PPG build up on initial work
by \cite{Zhan2020}. Each signal is described below and the visually depicted in \ref{fig:training-signals}

\textbf{Finger-PPG} is the ground truth signal collected simultaneously with
video recording using a contact finger sensor. This represents the standard
training signal for deep learning-based rPPG methods.

\textbf{Pseudo-PPG} is extracted directly from face videos using the
signal-processing-based method POS \cite{Wang2017}. Face detection and cropping
are performed as preprocessing steps, followed by POS signal extraction from the
cropped regions. The extracted signal is bandpass filtered (cutoff frequency: 0.7-3.0 Hz) using a
second-order Butterworth filter. Finally, the Hilbert transform is applied to
extract the signal envelope.

\textbf{Synced-PPG} is derived by temporally aligning Finger-PPG to Pseudo-PPG.
We compute the cross-correlation between the two signals and shift Finger-PPG to
maximize alignment. This preserves the morphological characteristics of Finger-PPG
while compensating for delays caused by the data collection setup and pulse transit
time (PTT) between face and finger measurement sites.

\subsection{Processing Pipeline}
\label{sec:pipeline}
The rPPG estimation pipeline consists of four stages. First, faces are detected
and extracted from input videos using YOLO5Face \cite{Qi2022}, following the
rPPG-Toolbox implementation \cite{Liu2023b}. Second, the deep learning models
(\autoref{sec:deep-learning-methods}) are trained to predict rPPG signals using
the training signals described in \autoref{sec:training-signals}. Third, predicted
signals are bandpass filtered (cutoff frequency: 0.75-3.0 Hz, corresponding to 45-180 BPM).
Finally, heart rate is estimated by identifying the peak frequency in the power spectral
density computed using Welch's method \cite{Welch1967a}.

%% file: sections/4_experiments.tex
\section{Experiments}

This section outlines the experimental protocols, datasets, implementation details, and evaluation metrics used to compare
the performance of different training signals.
We conduct two main evaluations: within-dataset cross-validation and cross-dataset generalization. Both are described below.

\subsection{Datasets}

We conducted experiments on CHILL \cite{Acharya2025} and PURE \cite{Stricker2014a} datasets,
chosen because they provide complementary challenges.
PURE includes substantial head movements under stable illumination, while CHILL features varying
illumination and physiological states (resting and elevated heart rate) with stationary
participants.
This diversity provides a comprehensive evaluation across different rPPG challenges.
Additionally, PURE uses hardware synchronization while CHILL uses software synchronization,
introducing different synchronization error characteristics.

The CHILL dataset \cite{Acharya2025} consists of 45 participants (28 females, 17 males). Data from
each participant was collected across 2 lighting settings and 2 heart rate conditions
(resting and elevated), resulting in 4 videos per participant. Videos were recorded
using a DSLR camera at 25 FPS, and ground truth signals were captured using contact
sensors at 1000 Hz.

The PURE dataset \cite{Stricker2014a} consists of 10 participants (2 females, 8 males). Each participant
was recorded in 6 different scenarios involving varying motion and head movements.
Videos were recorded at 30 FPS with frame-level timestamps, and finger PPG ground
truth was captured using a contact sensor at 1000 Hz. External hardware ensured
synchronization between video and PPG signals.

\begin{figure*}[ht]
	\centering
	\includegraphics[width=0.86\textwidth]{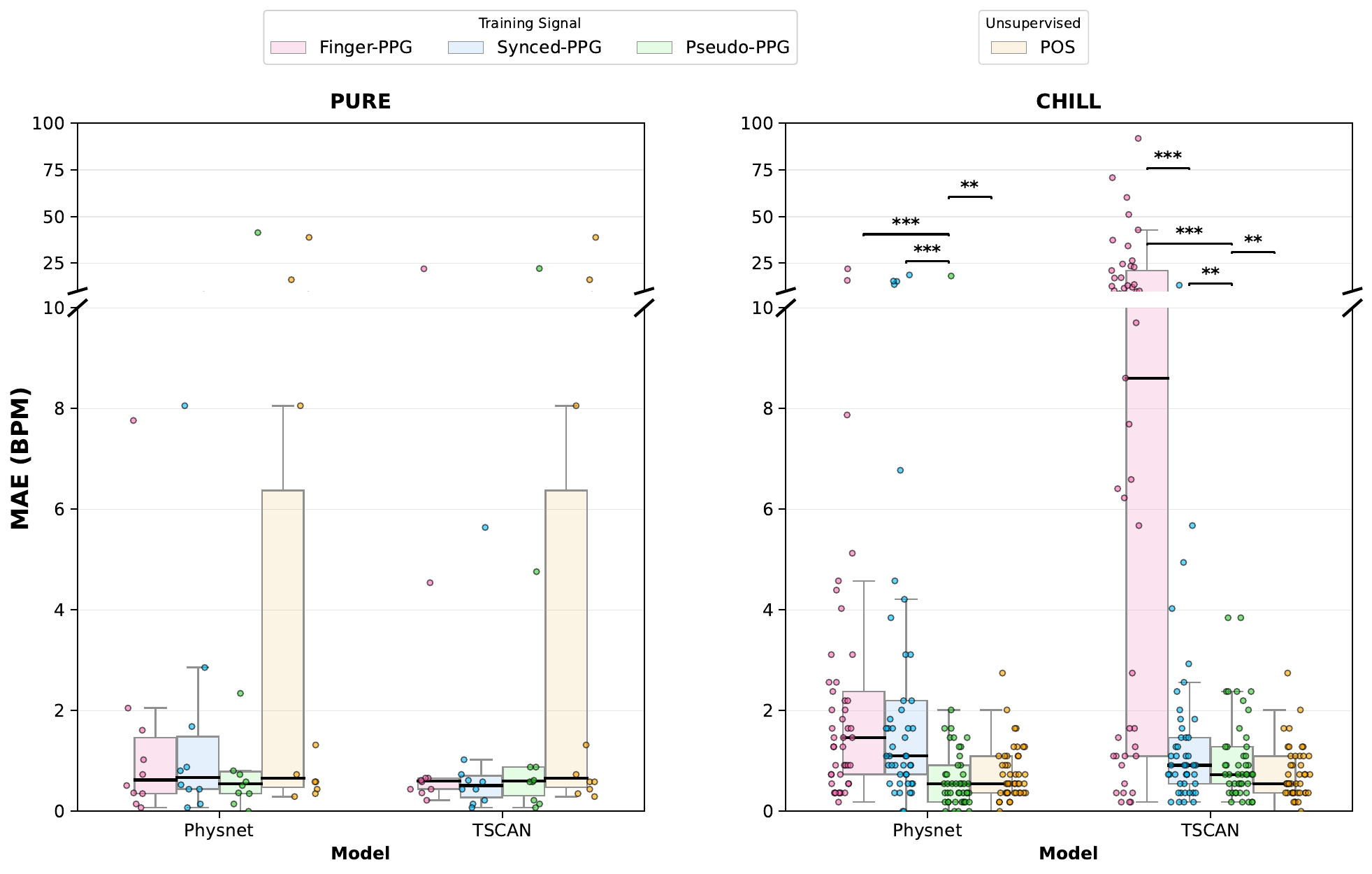}
	\caption{ \textbf{Distribution of Errors (MAE) of the within-dataset evaluations (\autoref{within-dataset}) for all rPPG methods.}
		Each point overlaid on the box plots represents the MAE for an individual participant for the corresponding
		training signal.
		Statistical significance between two conditions was assessed using the Wilcoxon signed-rank test. See  \autoref{tab:within-dataset-stats}
		for corresponding test statistics. We further report the baseline signal-processing-based method POS for comparison}
	\label{fig:experiment_1}
\end{figure*}

\begin{table*}[ht]
	\centering
	\caption{Wilcoxon Signed-Rank Test results for within-dataset comparisons, grouped by model and datasets used for training.
		For each comparison, the median error for each condition is reported along with the
		median of the paired differences ($\Delta$ Median), test statistic (W), P-value, and effect size (r).
		For the visualization see \autoref{fig:experiment_1} .
		*$p < 0.05$, **$p < 0.01$, ***$p < 0.001$}
	\label{tab:within-dataset-stats}
	\begin{tabular}{@{}llllrrrrrr@{}}
		\toprule
		Dataset                & Model                    & Comparison                & $n$ & Mdn$_1$ & Mdn$_2$ & $\Delta$ & $W$   & $r$  & $p$        \\
		\midrule
		\multirow{8}{*}{PURE}  & \multirow{4}{*}{Physnet} & Pseudo-PPG vs. Finger-PPG & 10  & 0.55    & 0.62    & 0.00     & 10.0  & -.56 & 1.000      \\
		                       &                          & Pseudo-PPG vs. Synced-PPG & 10  & 0.55    & 0.67    & -0.07    & 16.0  & -.37 & .279       \\
		                       &                          & Finger-PPG vs. Synced-PPG & 10  & 0.62    & 0.67    & -0.07    & 7.5   & -.64 & .172       \\
		                       &                          & Pseudo-PPG vs. POS        & 10  & 0.55    & 0.66    & -0.15    & 15.0  & -.40 & .410       \\
		\cmidrule(lr){2-10}
		                       & \multirow{4}{*}{TS-CAN}  & Pseudo-PPG vs. Finger-PPG & 10  & 0.60    & 0.60    & 0.07     & 15.0  & -.40 & .766       \\
		                       &                          & Pseudo-PPG vs. Synced-PPG & 10  & 0.60    & 0.51    & 0.07     & 5.0   & -.73 & .281       \\
		                       &                          & Finger-PPG vs. Synced-PPG & 10  & 0.60    & 0.51    & 0.04     & 8.0   & -.63 & .359       \\
		                       &                          & Pseudo-PPG vs. POS        & 10  & 0.60    & 0.66    & 0.04     & 26.5  & -.03 & .939       \\
		\midrule
		\multirow{8}{*}{CHILL} & \multirow{4}{*}{Physnet} & Pseudo-PPG vs. Finger-PPG & 45  & 0.55    & 1.46    & -0.73    & 84.5  & -.73 & $<$.001*** \\
		                       &                          & Pseudo-PPG vs. Synced-PPG & 45  & 0.55    & 1.10    & -0.55    & 17.0  & -.84 & $<$.001*** \\
		                       &                          & Finger-PPG vs. Synced-PPG & 45  & 1.46    & 1.10    & 0.00     & 399.0 & -.20 & .683       \\
		                       &                          & Pseudo-PPG vs. POS        & 45  & 0.55    & 0.55    & -0.18    & 148.0 & -.62 & .008**     \\
		\cmidrule(lr){2-10}
		                       & \multirow{4}{*}{TS-CAN}  & Pseudo-PPG vs. Finger-PPG & 45  & 0.73    & 8.61    & -7.14    & 46.5  & -.79 & $<$.001*** \\
		                       &                          & Pseudo-PPG vs. Synced-PPG & 45  & 0.73    & 0.92    & -0.18    & 91.5  & -.72 & .003**     \\
		                       &                          & Finger-PPG vs. Synced-PPG & 45  & 8.61    & 0.92    & 7.14     & 52.5  & -.78 & $<$.001*** \\
		                       &                          & Pseudo-PPG vs. POS        & 45  & 0.73    & 0.55    & 0.18     & 188.0 & -.55 & .003**     \\
		\bottomrule
	\end{tabular}
\end{table*}

\begin{figure*}[t]
	\centering
	\includegraphics[width=0.86\textwidth]{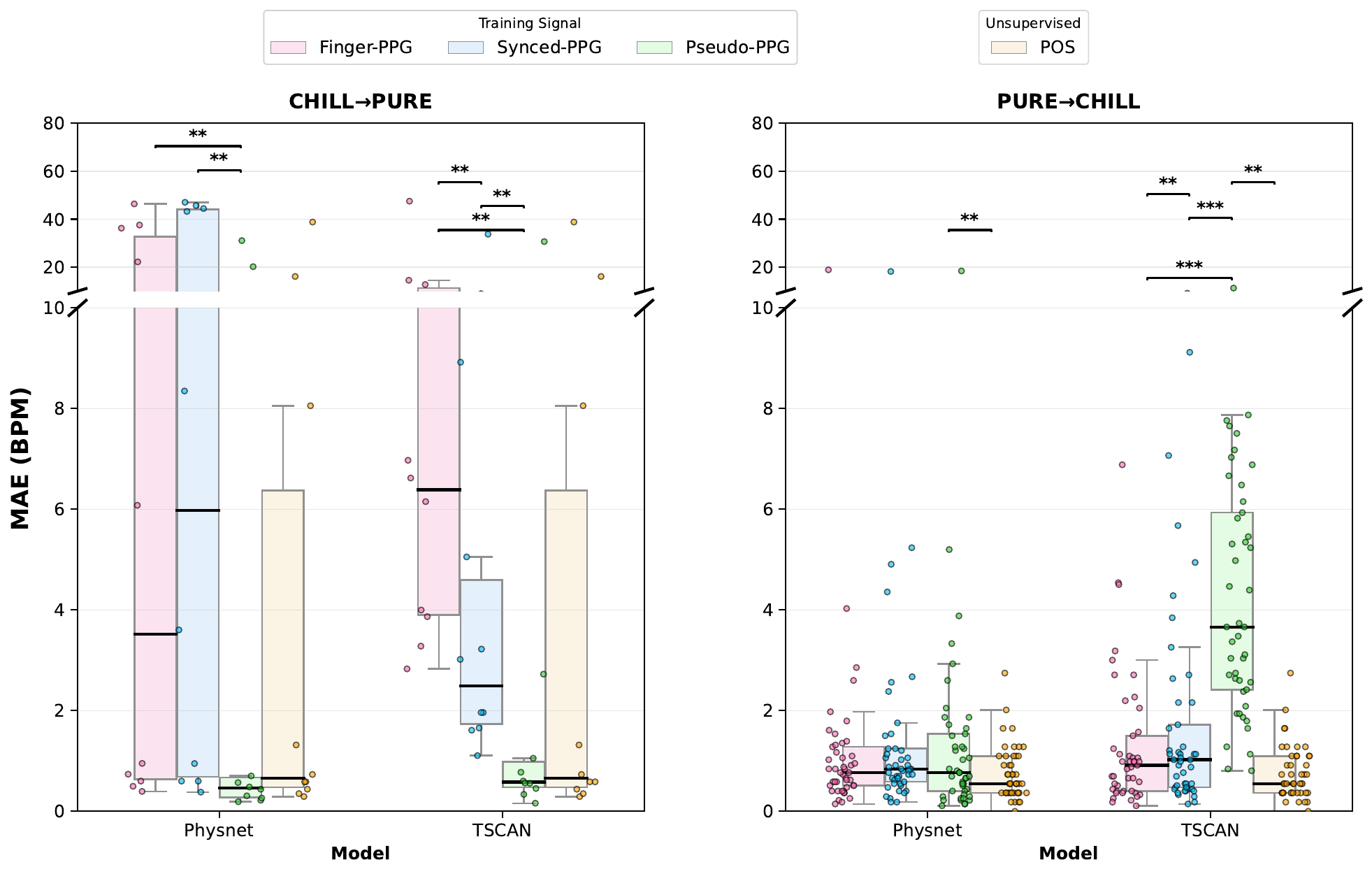}
	\caption{ \textbf{Distribution of Errors (MAE) of the cross-dataset evaluations (\autoref{cross-dataset}) for all rPPG methods.}
		Each point overlaid on the box plots represents the MAE for an individual participant for the corresponding
		training signal. Statistical significance between two conditions was assessed using the Wilcoxon signed-rank test. See  \autoref{tab:cross-dataset-stats} for the corresponding test statistics.}
	\label{fig:experiment_2}
\end{figure*}

\begin{table*}[!ht]
	\centering
	\caption{Wilcoxon Signed-Rank Test results for cross-dataset comparisons, grouped by model and datasets used for training and testing.
		For each comparison, the median error for each condition is reported along with the
		median of the paired differences ($\Delta$ Median), test statistic ($W$), $p$-value, and effect size ($r$).
		For the visualization see \autoref{fig:experiment_2}.
		*$p < 0.05$, **$p < 0.01$, ***$p < 0.001$}
	\label{tab:cross-dataset-stats}
	\begin{tabular}{@{}llllrrrrrr@{}}
		\toprule
		Train$\rightarrow$Test & Model                    & Comparison                & $n$ & Mdn$_1$ & Mdn$_2$ & $\Delta$ & $W$   & $r$     & $p$        \\
		\midrule
		\multirow{8}{*}{\shortstack[l]{CHILL$\rightarrow$PURE}}                                                                                           \\
		                       & \multirow{4}{*}{Physnet} & Pseudo-PPG vs. Finger-PPG & 10  & 0.46    & 3.52    & $-$3.10  & 2.0   & $-$0.82 & .006**     \\
		                       &                          & Pseudo-PPG vs. Synced-PPG & 10  & 0.46    & 5.98    & $-$5.56  & 1.0   & $-$0.85 & .004**     \\
		                       &                          & Finger-PPG vs. Synced-PPG & 10  & 3.52    & 5.98    & $-$0.44  & 6.0   & $-$0.69 & .055       \\
		                       &                          & Pseudo-PPG vs. POS        & 10  & 0.46    & 0.66    & $-$0.25  & 13.0  & $-$0.47 & .148       \\
		\cmidrule(lr){2-10}
		                       & \multirow{4}{*}{TS-CAN}  & Pseudo-PPG vs. Finger-PPG & 10  & 0.58    & 6.39    & $-$6.08  & 0.0   & $-$0.89 & .002**     \\
		                       &                          & Pseudo-PPG vs. Synced-PPG & 10  & 0.58    & 2.49    & $-$2.12  & 0.0   & $-$0.89 & .002**     \\
		                       &                          & Finger-PPG vs. Synced-PPG & 10  & 6.39    & 2.49    & 3.68     & 0.0   & $-$0.89 & .002**     \\
		                       &                          & Pseudo-PPG vs. POS        & 10  & 0.58    & 0.66    & 0.10     & 26.0  & $-$0.05 & .900       \\
		\midrule
		\multirow{8}{*}{\shortstack[l]{PURE$\rightarrow$CHILL}}                                                                                           \\
		                       & \multirow{4}{*}{Physnet} & Pseudo-PPG vs. Finger-PPG & 45  & 0.77    & 0.77    & $-$0.07  & 425.0 & $-$0.16 & .414       \\
		                       &                          & Pseudo-PPG vs. Synced-PPG & 45  & 0.77    & 0.84    & $-$0.07  & 365.5 & $-$0.26 & .194       \\
		                       &                          & Finger-PPG vs. Synced-PPG & 45  & 0.77    & 0.84    & 0.04     & 399.0 & $-$0.20 & .511       \\
		                       &                          & Pseudo-PPG vs. POS        & 45  & 0.77    & 0.55    & 0.22     & 273.5 & $-$0.41 & .006**     \\
		\cmidrule(lr){2-10}
		                       & \multirow{4}{*}{TS-CAN}  & Pseudo-PPG vs. Finger-PPG & 45  & 3.66    & 0.92    & 2.38     & 2.0   & $-$0.87 & $<$.001*** \\
		                       &                          & Pseudo-PPG vs. Synced-PPG & 45  & 3.66    & 1.03    & 2.56     & 57.5  & $-$0.77 & $<$.001*** \\
		                       &                          & Finger-PPG vs. Synced-PPG & 45  & 0.92    & 1.03    & $-$0.04  & 140.5 & $-$0.63 & .002**     \\
		                       &                          & Pseudo-PPG vs. POS        & 45  & 3.66    & 0.55    & 3.19     & 0.0   & $-$0.87 & $<$.001*** \\
		\bottomrule
	\end{tabular}
\end{table*}

\subsection{Evaluation Protocol}

\subsubsection{Within-Dataset Evaluation}
\label{within-dataset}
We perform 10-fold participant-level cross-validation on both datasets. Each
fold reserves 10\% of participants for testing and 90\% for training. The
training set is further split 90\%/10\% into train and validation subsets,
ensuring no participant overlap between sets. The validation set is used for
early stopping and selecting the best-performing model for test-set evaluation.

\subsubsection{Cross-Dataset Evaluation}
\label{cross-dataset}
Cross-dataset experiments use the entire source dataset for training (with
internal 90\%/10\% train-validation split) and the complete target dataset
for testing.
Multiple random seeds are employed for train-validation splits to assess stability.
Similar to within-dataset evaluation, the validation set is used for early stopping
and model selection.

\subsubsection{Evaluation Metric}
We compute MAE between predicted and ground truth heart rates:
\begin{equation}
	\text{MAE} = \frac{1}{N} \sum_{i=1}^{N} |HR_{\text{pred}}^{(i)} - HR_{\text{gt}}^{(i)}|
\end{equation}
where $N$ is the number of videos, $HR_{\text{pred}}$ is the heart
rate estimated from the predicted signal from the deep learning models,
and $HR_{\text{gt}}$ is the ground truth heart rate estimated from the contact device.

\subsubsection{Statistical Testing}
\label{statistical analysis}

Statistical analyses of heart rate estimation errors were performed using the
Wilcoxon signed-rank test, a non-parametric method suitable
for paired data.
To compare the influence of training signals on the performance, MAEs were calculated per participant on the entire dataset.
Our null hypothesis ($H_{0}$) was that the error distributions would not differ across training signals.
The alternative hypothesis ($H_{a}$) was that there would be difference across training signals.

\subsection{Implementation Details}

We implement two deep learning based methods using PyTorch Lightning \cite{Falcon2020},
adapted from rPPG-Toolbox \cite{Liu2023b}.
Code for reproducing the experiments is available at [anonymized for review].
Facial regions were detected using YOLO5Face \cite{Qi2022} with dynamic per-frame
detection.
After which, their respective preprocessing procedures described in \autoref{sec:deep-learning-methods} were applied.
The preprocessed videos were then chunked and rescaled. Physnet processes 128-frame chunks at 64×64 resolution, while TS-CAN processes 160-frame chunks at 36×36 resolution with a frame depth of 10.
We train all models for 30 epochs with batch size 64 using the SALSA optimizer
\cite{Kenneweg2024}, which eliminates the need for manual learning rate tuning and
ensures fair comparison across training signals.
All other hyperparameters used default values from the original implementations.
Mean squared error (MSE) of the waveform was used
as the loss function for both models across all three training signals.

%% file: sections/5_results.tex
\section{Results}

In this section we present the results of both the within-dataset evaluation (see \autoref{within-dataset})
and cross-dataset evaluations (see \autoref{cross-dataset}).

\subsubsection{Within-Dataset Results}
We first present the within-dataset (\autoref{within-dataset}) results.
We applied Wilcoxon signed-rank test to compare the performance between training signals (\autoref{sec:training-signals}) for Physnet and TS-CAN.
Quantitative results are presented in \autoref{tab:within-dataset-stats}
and visualized in \autoref{fig:experiment_1}.
On the CHILL dataset, training on Pseudo-PPG outperformed training on Finger-PPG for both Physnet and TS-CAN.
Similarly, training on Pseudo-PPG also outperformed Synced-PPG for Physnet and TS-CAN.
Training on Synced-PPG significantly outperformed Finger-PPG for TS-CAN, but not Physnet.
We further compared the difference between the performance of deep learning methods trained on Pseudo-PPG with signal-processing based method POS.
Both Physnet and TS-CAN had lower performance compared to POS.
On the PURE dataset, training signal choice (Finger-PPG, Synced-PPG, Pseudo-PPG) did not significantly affect performance for either architecture.
Similarly, Pseudo-PPG and POS showed no significant performance differences for either Physnet or TS-CAN.

\subsubsection{Cross-Dataset Results}
We now present cross-dataset \autoref{cross-dataset} results.
We applied Wilcoxon signed-rank test to compare the performance between training signals for Physnet and TS-CAN.
Quantitative results are presented in \autoref{tab:cross-dataset-stats}
and visualized in \autoref{fig:experiment_2}.

When trained on CHILL and tested on PURE, training on Pseudo-PPG significantly outperformed
training on Finger-PPG and Pseudo-PPG for both deep learning methods.
Training on Synced-PPG significantly outperformed Finger-PPG for TS-CAN, but not for for Physnet.
Additionally, there was no significant difference between training on Pseudo-PPG and the POS.

When training on PURE and testing on CHILL, the model performance varied based on the deep learning method.
For Physnet, training signal choice did not significantly affect performance.
However, for TS-CAN training on Pseudo-PPG significantly underperformed compared to Finger-PPG and Synced-PPG.
Training on Pseudo-PPG significantly underperformed POS for both the Physnet and TS-CAN.

\section{Sensitivity Analysis}
\label{sec:sensitivity-analysis}
\begin{figure}[!ht]
	\centering
	\includegraphics[width=\columnwidth]{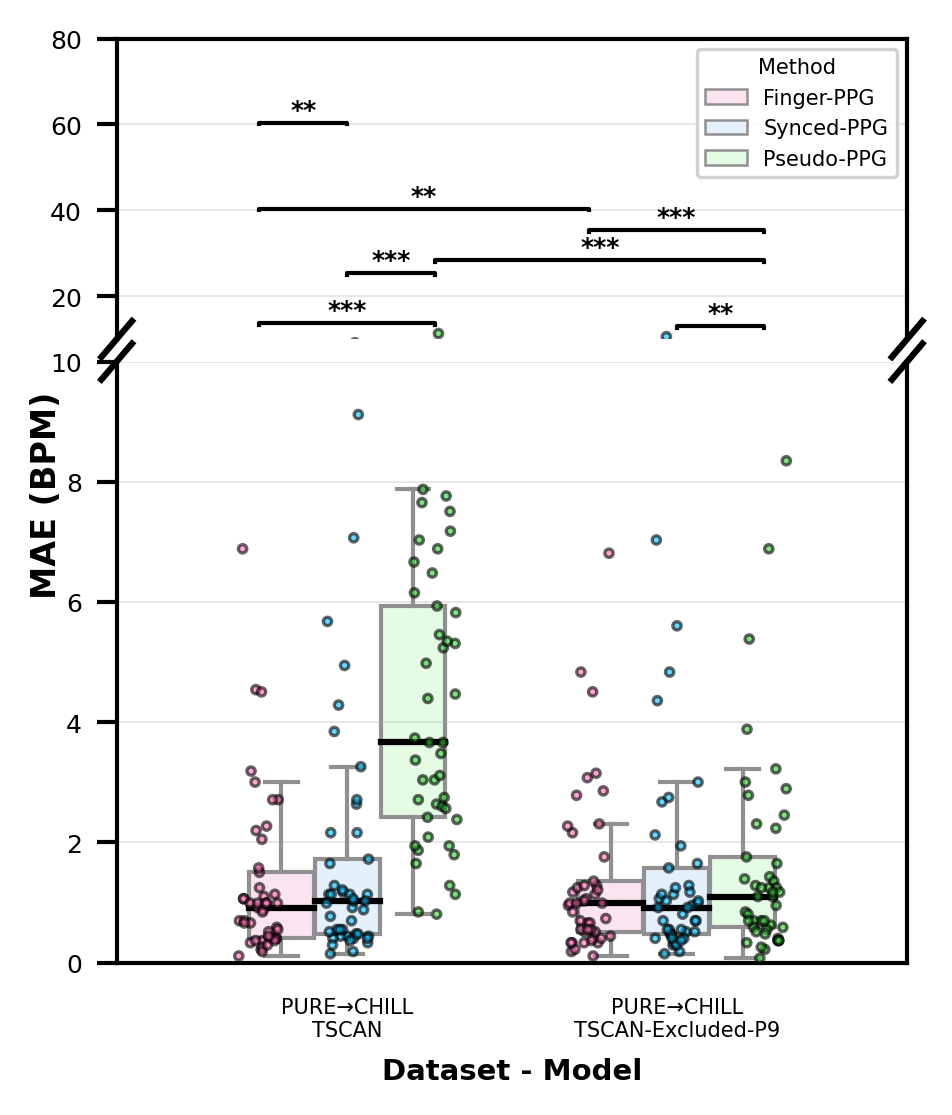}
	\caption{Participant-level MAE for TS-CAN sensitivity analysis  (\autoref{sec:sensitivity-analysis}) comparing three training signals  (Finger-PPG, Synced-PPG, Pseudo-PPG) on the full PURE dataset and with  participant 9 removed.}
	\label{fig:sensitivity}
\end{figure}

To assess TS-CAN sensitivity to training label quality, a sensitivity analysis was
performed on cross-dataset evaluation. Participant 9 from PURE exhibited substantially
higher POS estimation error than other participants, producing noisy Pseudo-PPG labels.
Both models were retrained on PURE with this participant excluded, then tested on CHILL
to determine whether removing low-quality labels improves generalization.
Wilcoxon signed-rank test was applied to compare the performance before and after participant exclusion, and to compare performance across training signals.
The results are reported in \autoref{tab:sensitivity-complete}
and visualized in \autoref{fig:sensitivity}.

Comparing the pairwise performance of before and after exclusion, we see that
excluding participant 9 from the PURE dataset significantly reduced the MAE for TS-CAN trained on Pseudo-PPG,
but significantly increased MAE for TS-CAN trained on Finger-PPG.
Training on Synced-PPG showed no significant difference in performance.
Comparing the performance of training on Pseudo-PPG and Finger-PPG after exclusion, we see that training on
Pseudo-PPG is significantly worse compared to training on Finger-PPG.

\begin{table*}[!ht]
	\centering
	\caption{Sensitivity analysis for cross-dataset evaluation (PURE→CHILL, n=45). Full represents training on the complete dataset. -P9 represents leaving out participant 9 from the training set.
		For each comparison, the median error under each condition is reported along with the
		median of the paired differences ($\Delta$ Median), test statistic ($W$), $p$-value, and effect size ($r$).
		For the visualization see \autoref{fig:sensitivity}.
		*$p < 0.05$, **$p < 0.01$, ***$p < 0.001$}
	\label{tab:sensitivity-complete}
	\begin{tabular}{@{}llrrrrrl@{}}
		\toprule
		Config & Comparison        & Mdn$_1$ & Mdn$_2$ & \textbf{$\Delta$} & \textbf{$W$} & \textbf{$r$} & \textbf{$p$} \\
		\midrule
		\multirow{3}{*}{-P9}
		       & Pseudo vs. Finger & 1.10    & 0.99    & 0.11              & 16.5         & $-$0.84      & $<$.001***   \\
		       & Pseudo vs. Synced & 1.10    & 0.92    & 0.15              & 218.5        & $-$0.50      & .002**       \\
		       & Finger vs. Synced & 0.99    & 0.92    & 0.00              & 375.5        & $-$0.24      & .838         \\
		\midrule
		\multirow{3}{*}{\shortstack[l]{Full                                                                             \\ vs. -P9}}
		       & Pseudo-PPG        & 3.66    & 1.10    & 2.31              & 15.0         & $-$0.85      & $<$.001***   \\
		       & Finger-PPG        & 0.92    & 0.99    & $-$0.04           & 160.5        & $-$0.60      & .004**       \\
		       & Synced-PPG        & 1.03    & 0.92    & 0.00              & 155.5        & $-$0.61      & .065         \\
		\bottomrule
	\end{tabular}
\end{table*}

%% file: sections/6_discussions.tex
\section{Discussion}

In this work, we systematically evaluated whether signal-processing-based rPPG can serve as
training signals for deep learning methods, replacing contact-based ground truth.
Our results demonstrate that Pseudo-PPG can substitute Finger-PPG under certain
conditions. We now examine these findings in detail to highlight important nuances.

Our experiments show that when datasets have imperfect synchronization between
video and contact sensors, Pseudo-PPG provides superior training signals compared
to poorly synchronized Finger-PPG. This finding aligns with \cite{Liu2023b},
who achieved strong generalization by training on POS-extracted labels (Pseudo-PPG)
for the BP4D+ dataset \cite{Zhang2016a}, which lacks accurate ground truth signals.

\cite{Braun2024a} showed that signal morphology and temporal alignment
from face-mounted sensors enable better-performing models. Our experiments reveal
similar morphological effects. On CHILL, Synced-PPG significantly outperforms
Finger-PPG, indicating that temporal alignment is crucial. However, Pseudo-PPG
outperforms Synced-PPG because its morphology more closely matches the signal
extracted from videos. This demonstrates the influence of both morphology and alignment
of the training signal has on the performance of the deep learning methods.

However, results from well-synchronized dataset reveal important nuances. On the one hand,
within-dataset experiments show that training signal choice does not significantly affect
performance, suggesting Pseudo-PPG as a viable alternative to Finger-PPG without the burden of collecting synchronized datasets.
On the other hand, cross-dataset experiments show TS-CAN fails to generalize when trained on Pseudo-PPG.
Our sensitivity analysis demonstrates this stems directly
from noisy Pseudo-PPG labels. Removing a single participant with
poor POS quality from the training set significantly improves generalization to a level.
In other words, the quality of the Pseudo-PPG directly influences model generalization.

Admittedly, assessing Pseudo-PPG label quality in the current framework still requires
ground truth, limiting scalability. However, developing methods to systematically assess
POS signal quality without ground truth would enable automatic training sample selection.
Further scaling rPPG methods to real-world deployments.

%% file: sections/7_conclusion.tex
\section{Conclusion}
We systematically evaluated whether signal processing extracted pseudo-labels can replace
contact sensor ground truth for training deep learning-based rPPG methods.
Our findings demonstrate that when well-synchronized ground truth is unavailable, pseudo-labels
provides superior training signals for models. Notably, high-quality pseudo-labels can even
match well-synchronized finger-PPG performance. This work establishes that signal-processing
based methods can generate valid training labels for deep learning models, enabling
large-scale rPPG development without dependence on labor-intensive synchronized data
collection.